\documentclass[10pt,journal,compsoc]{IEEEtran}
\usepackage{graphicx,times,amsmath,amssymb,url,subfigure,multirow}
\usepackage[nocompress]{cite}
\usepackage[lined,ruled,commentsnumbered]{algorithm2e}
 \allowdisplaybreaks \interdisplaylinepenalty=2500

\begin{document}

\title{Affect Estimation in 3D Space\\ Using Multi-Task Active Learning for Regression}
\author{Dongrui~Wu and Jian~Huang
\thanks{D.~Wu and J.~Huang are with the Key Laboratory of the Ministry of Education for Image Processing and Intelligent Control, School of Artificial Intelligence and Automation, Huazhong University of Science and Technology, Wuhan 430074, China. Email: drwu@hust.edu.cn, huang\_jan@hust.edu.cn.}}

\IEEEtitleabstractindextext{%
\begin{abstract}
Acquisition of labeled training samples for affective computing is usually costly and time-consuming, as affects are intrinsically subjective, subtle and uncertain, and hence multiple human assessors are needed to evaluate each affective sample. Particularly, for affect estimation in the 3D space of valence, arousal and dominance, each assessor has to perform the evaluations in three dimensions, which makes the labeling problem even more challenging. Many sophisticated machine learning approaches have been proposed to reduce the data labeling requirement in various other domains, but so far few have considered affective computing. This paper proposes two multi-task active learning for regression approaches, which select the most beneficial samples to label, by considering the three affect primitives simultaneously. Experimental results on the VAM corpus demonstrated that our optimal sample selection approaches can result in better estimation performance than random selection and several traditional single-task active learning approaches. Thus, they can help alleviate the data labeling problem in affective computing, i.e., better estimation performance can be obtained from fewer labeling queries.
\end{abstract}

}
\maketitle

\IEEEraisesectionheading{\section{Introduction}}

\IEEEPARstart{T}{he} amount of labeled training samples is critical to the performance of machine learning models. However, in many real-world applications it is easy to obtain unlabeled data, but labeling them may be very costly or time-consuming. This is particularly true for affective computing \cite{Picard1997}. Affects are very subjective, subtle, and uncertain. So, usually multiple human assessors are needed to obtain the groundtruth affect label for each affective sample (video, audio, image, text, etc.). For example, 14-16 assessors were used to evaluate each video clip in the DEAP dataset \cite{Koelstra2012}, and six to 17 assessors were used for each utterance in the VAM corpus \cite{Grimm2008}.

Many machine learning approaches have been proposed to alleviate the data labeling effort, including \cite{Harpale2012}:
\begin{enumerate}
\item \emph{Semi-supervised learning} \cite{Chapelle2006}, which uses typically a small amount of labeled data and a large amount of unlabeled data simultaneously in model training.
\item \emph{Transfer learning} \cite{Pan2010}, which makes use of data or knowledge from similar or relevant tasks to help the learning in a new task, which typical has a small number of labeled samples.
\item \emph{Multi-task learning} (MTL) \cite{Zhang2017}, in which multiple learning tasks are solved simultaneously, while exploiting commonalities and differences across them.
\item \emph{Active learning} \cite{Settles2009,drwuSAL2018,drwuRSVP2016}, which optimally selects the most informative unlabeled samples to label, so that a good learning model could be built from a small number of labeled samples.
\end{enumerate}
The above four approaches are independent and complementary, so they could be combined for even better performance. For example, we have developed a collaborative filtering approach, which integrates transfer learning and active class selection, a variant of active learning, for reducing the calibration data requirement in brain-computer interface (BCI).  We have also developed active weighted adaptation regularization \cite{drwuTNSRE2016}, which integrates active learning and domain adaptation, a specific form of transfer learning, for reducing the subject-specific calibration effort in BCI. Most recently, we have also developed an active semi-supervised transfer learning approach \cite{drwuSMC2017}, which integrates semi-supervised learning, transfer learning and active learning, for offline BCI calibration.

The focus of this paper is MTL, which has been successfully used in many real-world applications, including affective computing. For example, Jiang et al. \cite{Jiang2015a} proposed a multi-task fuzzy system that uses simultaneously independent sample information from each task and the inter-task common hidden structure among multiple tasks to enhance the generalization performance. It demonstrated promising performance in real-world applications including glutamic acid fermentation process modeling, polymer test plant modeling, wine preferences modeling, concrete slump modeling, etc. Su et al. \cite{Su2018} proposed MTL with low rank attribute embedding to perform person re-identification on multi-cameras, and demonstrated that it significantly outperformed existing single-task and multi-task approaches. Abadi et al. \cite{Abadi2014} proposed MTL-based regression models to simultaneously learn the relationship between low-level audio-visual features and high-level valence/arousal ratings from a collection of movie scenes. They can better predict valence and arousal ratings than scene-specific models. Xia and Liu \cite{Xia2017} integrated MTL and deep belief network to leverage activation and valence information for acoustic emotion recognition. Zhang et al. \cite{Zhang2017a} treated corpus, domain, and gender as different tasks in cross-corpus MTL, and showed that it outperformed approaches that treat the tasks as either identical or independent.

However, there have been very few approaches on integrating MTL and active learning. Reichart et al. \cite{Reichart2008} proposed multi-task active learning for linguistic annotations, which considers two annotation tasks (named entity and syntactic parse tree) and demonstrated promising performance. Zhang \cite{Zhang2010} studied multi-task active learning with output constraints, and demonstrated the effectiveness of the proposed framework in web information extraction and document classification. Li et al. \cite{Li2012b} proposed a multi-domain active learning approach for text classification, which jointly selects samples from multiple domains with duplicate information considered. Experiments on three real-world applications (sentiment classification, newsgroup classification and email spam filtering) showed that it outperformed several state-of-the-art single-task active learning approaches. Harpale \cite{Harpale2012} gave so far the most comprehensive study on multi-task active learning in his PhD Dissertation, which proposed approaches for homogeneous tasks, heterogeneous tasks, hierarchical classification, and collaborative filtering, and verified their performances in text classification, movie genre classification, image annotation, etc.

The above review shows that among the small number of studies on multi-task active learning, only one was related to affective computing (text sentiment classification), and none had considered regression problems\footnote{Chapter~6 of \cite{Harpale2012} presented an aspect model with Bayesian active learning algorithm for regression problems and applied it to two movie rating applications, but it is single-task active learning instead of multi-task active learning.}. However, affect estimation is a very pertinent application because affects intrinsically have multiple dimensions, e.g., affects can be represented in the 2D space of arousal and valence \cite{Russell1980}, or in the 3D space of arousal, valence, and dominance \cite{Mehrabian1980}. This paper fills the gap by proposing two multi-task active learning for regression (ALR) approaches, which extend two ALR approaches we proposed earlier \cite{drwuiGS2018} from single-task learning to MTL. Experimental results on the VAM corpus \cite{Grimm2008} demonstrate the effectiveness of our proposed approaches. Moreover, our proposed multi-task ALR approaches are generic, and they can also be used in other application domains beyond affective computing.

The remainder of this paper is organized as follows: Section~\ref{sect:MTALR} introduces two single-task ALR approaches based on greedy sampling, and their multi-task extensions. Section~\ref{sect:experiment} compares the performances of multi-task ALR with several state-of-the-art single-task ALR approaches on the VAM corpus. Section~\ref{sect:discussion} discusses why MTL should be preferred over single-task learning in affective computing. Section~\ref{sect:conclusions} draws conclusions and points out some future research directions.

\section{MT-ALR Using Greedy Sampling} \label{sect:MTALR}

This section extends two single-task ALR approaches we proposed recently \cite{drwuiGS2018} to MTL.

\subsection{Single-task GSy} \label{sect:GSy}

The GSy ALR approach, proposed in our recent work \cite{drwuiGS2018} for single-task regression, was inspired by the greedy sampling (GS) ALR approach proposed in \cite{Yu2010}. GS tries to select the most diverse samples in the \emph{input} space to label, whereas GSy aims to achieve diversity in the \emph{output} space.

The basic idea of GSy is as follows. Given a pool of unlabeled samples, GSy first selects a few samples using GS in the \emph{input} space to build an initial regression model, and then in each subsequent iteration selects a new sample located furthest away from all previously selected samples in the \emph{output} space to achieve diversity among the selected samples. Implementation details are given next.

Assume the pool consists of $N$ samples $\{\mathbf{x}_n\}_{n=1}^N$, initially none of which is labeled. Our goal is to select $K$ of them to label, and then construct an accurate regression model from them to estimate the outputs for the remaining $N-K$ samples. GSy selects the first sample as the one closest to the centroid of all $N$ samples (i.e., the one with the shortest average distance to the remaining $N-1$ samples), and the remaining $K-1$ samples incrementally.

To achieve diversity in the output space, GSy needs to know first the outputs (labels) of all samples, either true or estimated. Let $K_0$ be the minimum number of labeled samples required to build a regression model (in this paper $K_0$ is set as the number of features in the input space). GSy first uses GS in the input space to select the first $K_0$ samples to label. Without loss of generality, assume the first $k$ ($k<K_0$) samples have already been selected. For each of the remaining $N-k$ unlabeled samples $\{\mathbf{x}_n\}_{n=k+1}^N$, GSy computes first its distance to each of the $k$ labeled samples:
\begin{align}
d_{nm}^{\mathbf{x}}=||\mathbf{x}_n-\mathbf{x}_m||,\quad m=1,...,k; n=k+1,...,N \label{eq:dnmx}
\end{align}
then $d_n^{\mathbf{x}}$, the shortest distance from $\mathbf{x}_n$ to all $k$ labeled samples:
\begin{align}
d_n^{\mathbf{x}}=\min_m d_{nm}^{\mathbf{x}},\quad n=k+1,...,N \label{eq:dnx}
\end{align}
and finally selects the sample with the maximum $d_n^{\mathbf{x}}$ to label.

Once $K_0$ samples have been selected and labeled, a regression model can be constructed, and then GSy can select the remaining $K-K_0$ samples to achieve diversity in the output space. Without loss of generality, assume the first $k$ ($k\ge K_0$) samples have already been labeled with outputs $\{y_m\}_{m=1}^k$, and a regression model $f(\mathbf{x})$ has been constructed. For each of the remaining $N-k$ unlabeled samples $\{\mathbf{x}_n\}_{n=k+1}^N$, GSy computes first its distance to each of the $k$ outputs:
\begin{align}
d_{nm}^y=||f(\mathbf{x}_n)-y_m||,\quad m=1,...,k; n=k+1,...,N \label{eq:dnmy}
\end{align}
and $d_n^y$, the shortest distance from $f(\mathbf{x}_n)$ to $\{y_m\}_{m=1}^k$:
\begin{align}
d_n^y=\min_m d_{nm}^y,\quad n=k+1,...,N \label{eq:dny}
\end{align}
and then selects the sample with the maximum $d_n^y$ to label.

For the completeness of this paper, the pseudo-code of single-task GSy, initially proposed in \cite{drwuiGS2018}, is shown in Algorithm~\ref{alg:GSy}.

\begin{algorithm}[h] %\DontPrintSemicolon
\KwIn{$N$ unlabeled samples, $\{\mathbf{x}_n\}_{n=1}^N$\;
\hspace*{10mm} $K$, the maximum number of labels to query.}
\KwOut{The regression model $f(\mathbf{x})$.}
\tcp{Initialize the first selection}
Set $Z=\{\mathbf{x}_n\}_{n=1}^N$, and $S=\emptyset$\;
Identify $\mathbf{x}'$, the sample closest to the centroid of $Z$\;
Move $\mathbf{x}'$ from $Z$ to $S$\;
Re-index the sample in $S$ as $\mathbf{x}_1$, and the samples in $Z$ as $\{\mathbf{x}_n\}_{n=2}^N$\;
\tcp{Select $K_0-1$ more samples incrementally using GS in the input space}
Identify $K_0$, the minimum number of labeled samples required to construct $f(\mathbf{x})$\;
\For{$k=1,...,K_0-1$}{
\For{$n=k,...,N$}{
Compute $d_n^{\mathbf{x}}$ in (\ref{eq:dnx})\;}
Identify the $\mathbf{x}'$ that has the largest $d_n^{\mathbf{x}}$\;
Move $\mathbf{x}'$ from $Z$ to $S$\;
Re-index the samples in $S$ as $\{\mathbf{x}_m\}_{m=1}^{k+1}$, and the samples in $Z$ as $\{\mathbf{x}_n\}_{n=k+2}^N$\;}
Query to label the $K_0$ samples in $S$\;
Construct the regression model $f(\mathbf{x})$ from $S$\;
\tcp{Select $K-K_0$ more samples incrementally}
\For{$k=K_0,...,K-1$}{
\For{$n=k,...,N$}{
Compute $d_n^y$ in (\ref{eq:dny})\;}
Identify the $\mathbf{x}'$ that has the largest $d_n^y$\;
Move $\mathbf{x}'$ from $Z$ to $S$\;
Query to label $\mathbf{x}'$ in $S$\;
Re-index the samples in $S$ as $\{\mathbf{x}_m\}_{m=1}^{k+1}$, and the samples in $Z$ as $\{\mathbf{x}_n\}_{n=k+2}^N$\;
Update the regression model $f(\mathbf{x})$ using $S$.}
\caption{GSy for single-task ALR \cite{drwuiGS2018}.} \label{alg:GSy}
\end{algorithm}

\subsection{Multi-Task GSy (MT-GSy)} \label{sect:MT-GSy}

The original GSy was proposed for single-task learning, i.e., each sample in the input space has only one output (task). This subsection extends it to MTL.

Let $P$ be the number of tasks (dimensionality of the output space), i.e., $\mathbf{y}_n=(y_{n,1},...,y_{n,P})^T$. Multi-task GSy (MT-GSy) tries to select samples can benefit all $P$ tasks simultaneously.

MT-GSy first uses GSy to select and label $K_0$ samples. It then builds $P$ regression models $\{f_p(\mathbf{x})\}_{p=1}^P$, for the $P$ tasks, and next selects the remaining $K-K_0$ samples to achieve diversity in the $P$ output spaces simultaneously. Without loss of generality, assume the first $k$ ($k\ge K_0$) samples have already been labeled with outputs $\{\mathbf{y}_m\}_{m=1}^k$, and $P$ regression models $f_p(\mathbf{x})$ ($p=1,...,P$) have been built. For each of the remaining $N-k$ unlabeled samples $\{\mathbf{x}_n\}_{n=k+1}^N$, MT-GSy computes first its distance to each of the $k$ outputs, for each of the $P$ tasks:
\begin{align}
d_{nm,p}^\mathbf{y}=||f_p(\mathbf{x}_n)-y_m||  \label{eq:dnmyp}
\end{align}
where $m=1,...,k$, $n=k+1,...,N$, and $p=1,...,P$. MT-GSy then computes $d_n^\mathbf{y}$, the product of the shortest distances from $f_p(\mathbf{x}_n)$ to $\{y_{m,p}\}_{m=1}^k$, $p=1,...,P$:
\begin{align}
d_n^\mathbf{y}=\min_m \prod_{p=1}^P d_{nm,p}^\mathbf{y},\quad n=k+1,...,N \label{eq:dnyp}
\end{align}
and selects the sample with the maximum $d_n^\mathbf{y}$ to label. The pseudo-code of MT-GSy is given in Algorithm~\ref{alg:MT-GSy}.

It's interesting to note that:
\begin{enumerate}
\item In (\ref{eq:dnyp}) we combine the $P$ $d_{nm,p}^\mathbf{y}$ using product instead of summation, to avoid the problem that different task outputs may have different scales, and hence a task with large outputs may dominate other tasks.
\item MT-GSy degrades to the single-task GSy when $P=1$.
\item We considered a simple multi-task setting that all tasks share the same inputs. More general settings that allow different tasks to have different inputs will be considered in our future research.
\end{enumerate}

\begin{algorithm}[h] %\DontPrintSemicolon
\KwIn{$N$ unlabeled samples, $\{\mathbf{x}_n\}_{n=1}^N$\;
\hspace*{10mm} $K$, the maximum number of labels to query.}
\KwOut{$P$ regression models $f_p(\mathbf{x})$, $p=1,...,P$.}
\tcp{Initialize the first selection}
Set $Z=\{\mathbf{x}_n\}_{n=1}^N$, and $S=\emptyset$\;
Identify $\mathbf{x}'$, the sample closest to the centroid of $Z$\;
Move $\mathbf{x}'$ from $Z$ to $S$\;
Re-index the sample in $S$ as $\mathbf{x}_1$, and the samples in $Z$ as $\{\mathbf{x}_n\}_{n=2}^N$\;
\tcp{Select $K_0-1$ more samples incrementally using GS in the input space}
Identify $K_0$, the minimum number of labeled samples required to construct $f(\mathbf{x})$\;
\For{$k=1,...,K_0-1$}{
\For{$n=k,...,N$}{
Compute $d_n^{\mathbf{x}}$ in (\ref{eq:dnx})\;}
Identify the $\mathbf{x}'$ that has the largest $d_n^{\mathbf{x}}$\;
Move $\mathbf{x}'$ from $Z$ to $S$\;
Re-index the samples in $S$ as $\{\mathbf{x}_m\}_{m=1}^{k+1}$, and the samples in $Z$ as $\{\mathbf{x}_n\}_{n=k+2}^N$\;}
Query to label the $K_0$ samples in $S$, for all $P$ tasks\;
Construct $P$ regression models $f_p(\mathbf{x})$ from $S$, one for each task\;
\tcp{Select $K-K_0$ more samples incrementally}
\For{$k=K_0,...,K-1$}{
\For{$n=k,...,N$}{
Compute $d_n^\mathbf{y}$ in (\ref{eq:dnyp})\;}
Identify the $\mathbf{x}'$ that has the largest $d_n^\mathbf{y}$\;
Move $\mathbf{x}'$ from $Z$ to $S$\;
Query to label $\mathbf{x}'$ in $S$, for all $P$ tasks\;
Re-index the samples in $S$ as $\{\mathbf{x}_m\}_{m=1}^{k+1}$, and the samples in $Z$ as $\{\mathbf{x}_n\}_{n=k+2}^N$\;
Update the $P$ regression models $f_p(\mathbf{x})$ using $S$.}
\caption{MT-GSy for multi-task ALR.} \label{alg:MT-GSy}
\end{algorithm}

\subsection{Single-Task iGS} \label{sect:iGS}

The improved greedy sampling (iGS) approach, proposed in our recent work \cite{drwuiGS2018} for single-task learning, considers the diversity in both the input and output spaces.

Like the single-task GSy, initially the pool consists of $N$ unlabeled samples and zero labeled sample. In iGS we again set $K_0$ to be the number of features in the input space, and use GS in the input space to select the first $K_0$ samples to label. Assume the first $k$ samples have already been labeled with labels $\{y_n\}_{n=1}^k$. For each of the remaining $N-k$ unlabeled sample $\{\mathbf{x}_n\}_{n=k+1}^N$, iGS computes first its distance to each of the $k$ labeled samples in the input space:
\begin{align}
d_{nm}^{\mathbf{x}}=||\mathbf{x}_n-\mathbf{x}_m||,\quad m=1,...,k; n=k+1,...,N \label{eq:dnmx}
\end{align}
and $d_{nm}^y$ in (\ref{eq:dnmy}), and then $d_n^{\mathbf{x}y}$:
\begin{align}
d_n^{\mathbf{x}y}=\min_m d_{nm}^{\mathbf{x}}d_{nm}^y,\quad n=k+1,...,N \label{eq:dnxy}
\end{align}
Next, iGS selects the sample with the maximum $d_n^{\mathbf{x}y}$ to label.

In summary, iGS uses the same procedure as GSy to select the first $K_0$ samples to build an initial regression model, and then in each subsequent iteration a new sample located furthest away from all previously selected samples in both the input and output spaces is selected to achieve balanced diversity among the selected samples. Its pseudo-code, originally proposed in \cite{drwuiGS2018}, is given in Algorithm~\ref{alg:iGS}, for the completeness of this paper.

\begin{algorithm}[h] %\DontPrintSemicolon
\KwIn{$N$ unlabeled samples, $\{\mathbf{x}_n\}_{n=1}^N$\;
\hspace*{10mm} $K$, the maximum number of labels to query.}
\KwOut{The regression model $f(\mathbf{x})$.}
\tcp{Initialize the first selection}
Set $Z=\{\mathbf{x}_n\}_{n=1}^N$, and $S=\emptyset$\;
Identify $\mathbf{x}'$, the sample closest to the centroid of $Z$\;
Move $\mathbf{x}'$ from $Z$ to $S$\;
Re-index the sample in $S$ as $\mathbf{x}_1$, and the samples in $Z$ as $\{\mathbf{x}_n\}_{n=2}^N$\;
\tcp{Select $K_0-1$ more samples incrementally using GS in the input space}
Identify $K_0$, the minimum number of labeled samples required to construct $f(\mathbf{x})$\;
\For{$k=1,...,K_0-1$}{
\For{$n=k,...,N$}{
Compute $d_n^{\mathbf{x}}$ in (\ref{eq:dnx})\;}
Identify the $\mathbf{x}'$ that has the largest $d_n^{\mathbf{x}}$\;
Move $\mathbf{x}'$ from $Z$ to $S$\;
Re-index the samples in $S$ as $\{\mathbf{x}_m\}_{m=1}^{k+1}$, and the samples in $Z$ as $\{\mathbf{x}_n\}_{n=k+2}^N$\;}
Query to label the $K_0$ samples in $S$\;
Construct the regression model $f(\mathbf{x})$ from $S$\;
\tcp{Select $K-K_0$ more samples incrementally}
\For{$k=K_0,...,K-1$}{
\For{$n=k,...,N$}{
Compute $d_n^{\mathbf{x}y}$ in (\ref{eq:dnxy})\;}
Identify the $\mathbf{x}'$ that has the largest $d_n^{\mathbf{x}y}$\;
Move $\mathbf{x}'$ from $Z$ to $S$\;
Query to label $\mathbf{x}'$ in $S$\;
Re-index the samples in $S$ as $\{\mathbf{x}_m\}_{m=1}^{k+1}$, and the samples in $Z$ as $\{\mathbf{x}_n\}_{n=k+2}^N$\;
Update the regression model $f(\mathbf{x})$ using $S$.}
\caption{iGS for single-task ALR \cite{drwuiGS2018}.} \label{alg:iGS}
\end{algorithm}

\subsection{Multi-Task iGS (MT-iGS)} \label{sect:MT-iGS}

This subsection extends the single-task iGS to multi-task iGS (MT-iGS). Similar to MT-GSy, here we again consider a simple multi-task setting that all $P$ tasks share the same inputs.

MT-iGS first uses iGS to select and label the $K_0$ samples. It then builds $P$ regression models $\{f_p(\mathbf{x})\}_{p=1}^P$ for the $P$ tasks, and next selects the remaining $K-K_0$ samples to achieve diversity in both the input and output spaces. Without loss of generality, assume the first $k$ ($k\ge K_0$) samples have already been labeled with outputs $\{\mathbf{y}_m\}_{m=1}^k$, and $P$ regression models $f_p(\mathbf{x})$ ($p=1,...,P$) have been built. For each of the remaining $N-k$ unlabeled samples $\{\mathbf{x}_n\}_{n=k+1}^N$, MT-iGS computes $d_{nm}^{\mathbf{x}}$ in (\ref{eq:dnmx}) and $d_{nm,p}^\mathbf{y}$ in (\ref{eq:dnmyp}), and then $d_n^\mathbf{xy}$:
\begin{align}
d_n^\mathbf{xy}=\min_m d_{nm}^{\mathbf{x}}\prod_{p=1}^P d_{nm,p}^\mathbf{y},\quad n=k+1,...,N \label{eq:dnxyp}
\end{align}
and selects the sample with the maximum $d_n^\mathbf{xy}$ to label. The pseudo-code of MT-iGS is given in Algorithm~\ref{alg:MT-iGS}.

\begin{algorithm}[h] %\DontPrintSemicolon
\KwIn{$N$ unlabeled samples, $\{\mathbf{x}_n\}_{n=1}^N$\;
\hspace*{10mm} $K$, the maximum number of labels to query.}
\KwOut{$P$ regression models $f_p(\mathbf{x})$, $p=1,...,P$.}
\tcp{Initialize the first selection}
Set $Z=\{\mathbf{x}_n\}_{n=1}^N$, and $S=\emptyset$\;
Identify $\mathbf{x}'$, the sample closest to the centroid of $Z$\;
Move $\mathbf{x}'$ from $Z$ to $S$\;
Re-index the sample in $S$ as $\mathbf{x}_1$, and the samples in $Z$ as $\{\mathbf{x}_n\}_{n=2}^N$\;
\tcp{Select $K_0-1$ more samples incrementally using GS in the input space}
Identify $K_0$, the minimum number of labeled samples required to construct $f(\mathbf{x})$\;
\For{$k=1,...,K_0-1$}{
\For{$n=k,...,N$}{
Compute $d_n^{\mathbf{x}}$ in (\ref{eq:dnx})\;}
Identify the $\mathbf{x}'$ that has the largest $d_n^{\mathbf{x}}$\;
Move $\mathbf{x}'$ from $Z$ to $S$\;
Re-index the samples in $S$ as $\{\mathbf{x}_m\}_{m=1}^{k+1}$, and the samples in $Z$ as $\{\mathbf{x}_n\}_{n=k+2}^N$\;}
Query to label the $K_0$ samples in $S$, for all $P$ tasks\;
Construct $P$ regression models $f_p(\mathbf{x})$ from $S$, one for each task\;
\tcp{Select $K-K_0$ more samples incrementally}
\For{$k=K_0,...,K-1$}{
\For{$n=k,...,N$}{
Compute $d_n^{\mathbf{xy}}$ in (\ref{eq:dnxyp})\;}
Identify the $\mathbf{x}'$ that has the largest $d_n^{\mathbf{x}y}$\;
Move $\mathbf{x}'$ from $Z$ to $S$\;
Query to label $\mathbf{x}'$ in $S$, for all $P$ tasks\;
Re-index the samples in $S$ as $\{\mathbf{x}_m\}_{m=1}^{k+1}$, and the samples in $Z$ as $\{\mathbf{x}_n\}_{n=k+2}^N$\;
Update the $P$ regression models $f_p(\mathbf{x})$ using $S$.}
\caption{MT-iGS for multi-task ALR.} \label{alg:MT-iGS}
\end{algorithm}

Similar to MT-GSy, we can also note that:
\begin{enumerate}
\item In (\ref{eq:dnxyp}) we combine $d_{nm}^\mathbf{x}$ and the $P$ $d_{nm,p}^\mathbf{y}$ using product instead of summation, to avoid the problem that the inputs and different task outputs may have different scales, and hence one distance may dominate others.
\item MT-iGS degrades to the single-task iGS when $P=1$.
\end{enumerate}

\section{Experiment} \label{sect:experiment}

The VAM corpus \cite{Grimm2008} is used in this section to demonstrate the performances of MT-GSy and MT-iGS.

\subsection{Dataset and Feature Extraction}

The VAM corpus was released in ICME2008 \cite{Grimm2008} and has been used in many studies \cite{Grimm2007a,Grimm2007b,drwuICME2010,drwuInterSpeech2010}. It contains spontaneous speech with authentic emotions recorded from guests in a German TV talk-show \emph{Vera am Mittag} (\emph{Vera at Noon} in English). There are 947 emotional utterances from 47 speakers (11m/36f).  Each sentence was evaluated by 6-17 listeners in the 3D space of valence, arousal and dominance, and the evaluations were merged by a weighted average to obtain the groundtruth emotion primitives in $[-1,1]$ \cite{Grimm2005b}.

The same $46$ acoustic features extracted in our previous research \cite{drwuICME2010,drwuInterSpeech2010} were used again in this paper. They included nine pitch features, five duration features, six energy features, and 26 Mel Frequency Cepstral Coefficient (MFCC) features. Each feature was then normalized to mean 0 and standard deviation 1.

\subsection{Sample Selection Algorithms}

We compared the performances of nine sample selection algorithms:
\begin{enumerate}
\item Baseline 1 (BL1), which randomly selects all $K$ samples.
\item Baseline 2 (BL2), which assumes all samples in the training pool are labeled, and uses them to build a regression model. BL2 represents the upper bound of the performance we could get given a specific training pool.
\item Expected model change maximization (EMCM) \cite{Cai2013}, which selects the sample with the maximum expected model change to label. EMCM is for single-task learning.
\item Query-by-Committee (QBC) \cite{RayChaudhuri1995}, which selects the sample with the maximum variance (computed from a committee of regression models) to label. QBC is for single-task learning.
\item GSx, which was introduced in our recent research \cite{drwuiGS2018}. It is almost identical to the GS approach in \cite{Yu2010}, except that the first sample is selected as the one closest to the centroid of all $N$ unlabeled samples. Since GSx considers only the diversity in the input space, and in this paper all $P$ tasks share the same input space, it can be used in both single-task and MTL settings, without any modification.
\item GSy, which has been introduced in Section~\ref{sect:GSy}.
\item MT-GSy, which has been introduced in Section~\ref{sect:MT-GSy}.
\item iGS, which has been introduced in Section~\ref{sect:iGS}.
\item MT-iGS, which has been introduced in Section~\ref{sect:MT-iGS}.
\end{enumerate}

\subsection{Performance Evaluation Process} \label{sect:process}

For the 947 samples in the VAM corpus, we first randomly selected 30\% as the training pool and the remaining 70\% as the test dataset, initialized the first $K_0$ labeled samples ($K_0$ is the dimensionality of the input space) either randomly (for BL, QBC and EMCM) or by GSy (for GSx, GSy, iGS, MT-GSy, and MT-iGS), identified one sample to label in each iteration by different algorithms, built linear regression models, and computed the root mean squared error (RMSE) and correlation coefficient (CC) as the performance measures on the test dataset. The iteration terminated when all samples in the training pool were selected.

To obtain statistically meaningful results, we ran this evaluation process 100 times for each algorithm, each time with a randomly chosen training pool containing 30\% unlabeled samples.

\subsection{Experimental Results} \label{sect:RR}

First, ridge regression (RR) was used as the linear regression model, and $\lambda=10/K$ was used in its objective function $\underset{\boldsymbol{\beta}}{\operatorname{min}} (\| \mathbf{y}-\mathbf{X} \boldsymbol{\beta} \|^2 + \lambda \|\boldsymbol{\beta}\|^2)$. Given a fixed training pool, GSx will select a fixed sequence of samples to label because it only considers the diversity in the input space, regardless of how many tasks are there. MT-GSy (MT-iGS) also generates a fixed sequence of samples to label because it always considers all tasks simultaneously. However, each single-task ALR approach (EMCM, QBC, GSy and iGS) will give a different sequence of samples when a different task is considered. So, we compare the performances of the sample selection algorithms under three scenarios: 1) \emph{Valence} estimation is considered in the single-task ALR approaches; 2) \emph{Arousal} estimation is considered in the single-task ALR approaches; and, 3) \emph{Dominance} estimation is considered in the single-task ALR approaches.

\begin{figure*}[htpb]\centering
\includegraphics[width=.95\linewidth,clip]{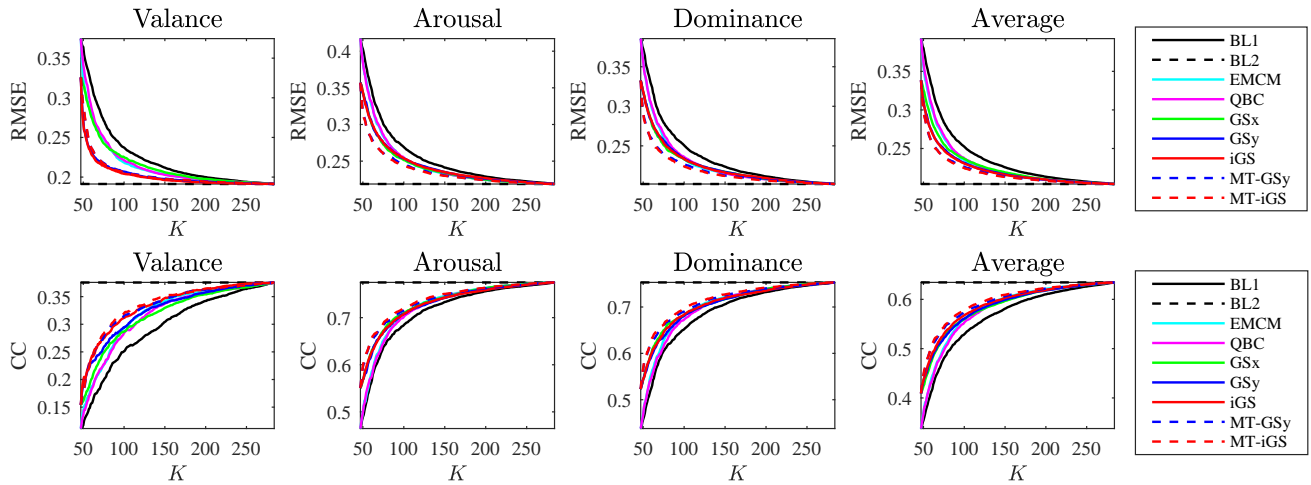}
\caption{Performances of the sample selection algorithms, averaged over 100 runs, when the single-task ALR approaches focused on \emph{Valence} estimation. The last column shows the average RMSE and CC across the three tasks. RR was used as the regression model.} \label{fig:RR1}
\end{figure*}

\begin{figure*}[htpb]\centering
\includegraphics[width=.95\linewidth,clip]{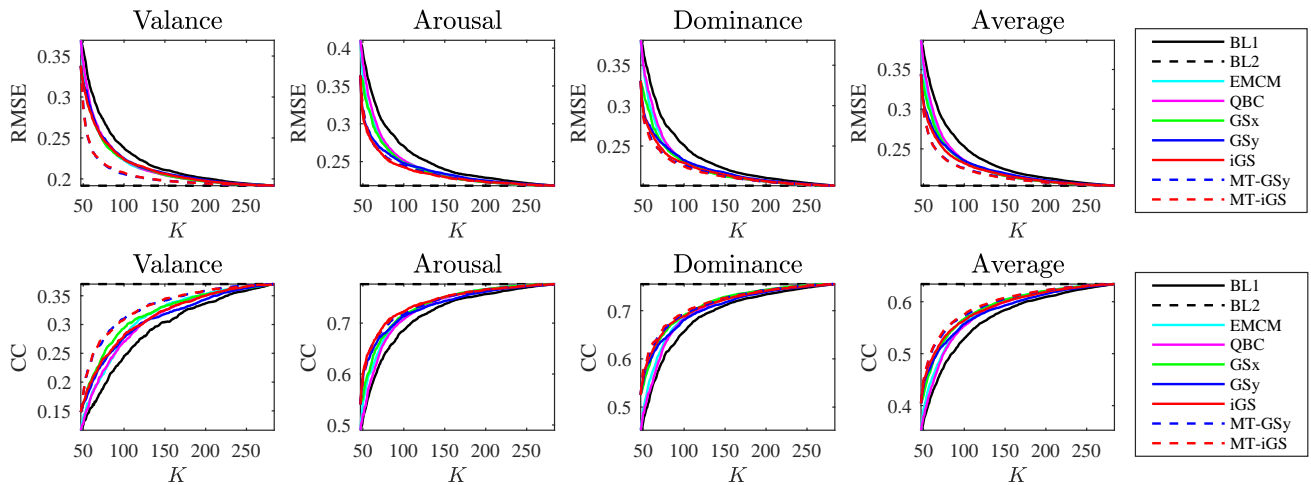}
\caption{Performances of the sample selection algorithms, averaged over 100 runs, when the single-task ALR approaches focused on \emph{Arousal} estimation. The last column shows the average RMSE and CC across the three tasks. RR was used as the regression model.} \label{fig:RR2}
\end{figure*}

\begin{figure*}[htpb]\centering
\includegraphics[width=.95\linewidth,clip]{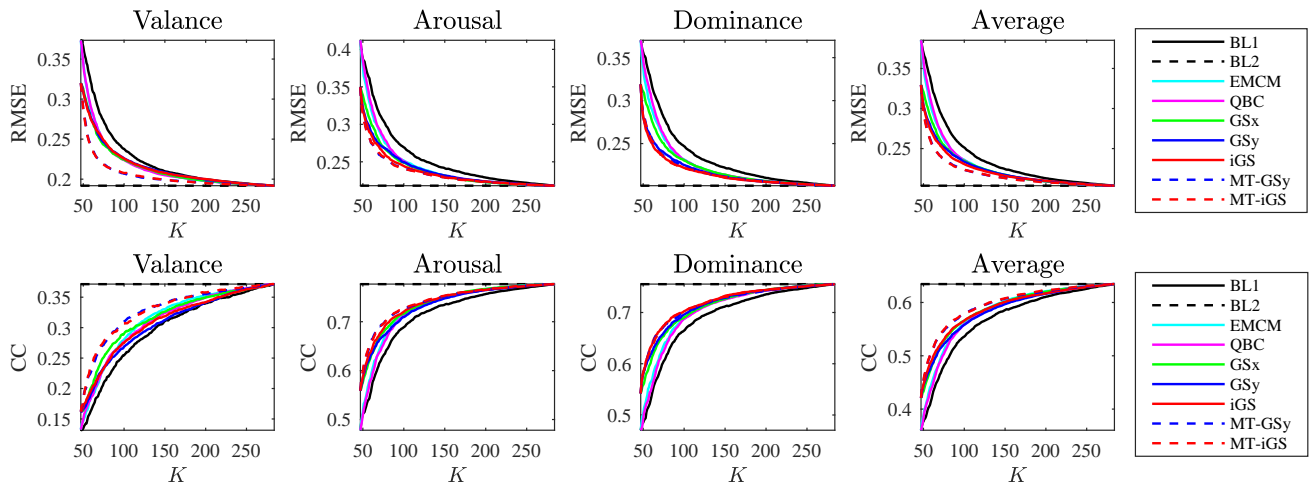}
\caption{Performances of the sample selection algorithms, averaged over 100 runs, when the single-task ALR approaches focused on \emph{Dominance} estimation. The last column shows the average RMSE and CC across the three tasks. RR was used as the regression model.} \label{fig:RR3}
\end{figure*}

The results are shown in Figs.~\ref{fig:RR1}-\ref{fig:RR3}, respectively, where the RMSEs and CCs have been averaged over 100 runs. In each subfigure, the first column shows the results when the single-task ALR approaches focused on Valence estimation, the second column on Arousal estimation, and the third column on Dominance estimation. The last column shows the average across the first three columns.

Observe from Fig.~\ref{fig:RR1} that:
\begin{enumerate}
\item Generally as $K$ increased, all eight sample selection algorithms (excluding BL2, which did not change with $K$) achieved better performance (smaller RMSE and larger CC), which is intuitive, because more labeled training samples generally result in a more reliable RR model.
\item When $K=K_0=46$ (the first point in each subfigure of Fig.~\ref{fig:RR1}), the five GS based ALR approaches (GSx, GSy, iGS, MT-GSy and MT-iGS), which initialized the $K_0$ samples by considering the diversity in the input space, all had better performances than the other three approaches (BL1, EMCM and QBC), which initialized the $K_0$ samples randomly.
\item For Valence estimation, which was the task that all single-task ALR approaches focused on, GSy and iGS achieved comparable performance as MT-GSy and MT-iGS; however, for the other two tasks (Arousal and Dominance estimations), MT-GSy and MT-iGS achieved better performances.
\item When all three tasks are considered together (the last column of Fig.~\ref{fig:RR1}), on average all ALR approaches outperformed BL1, all GS based approaches outperformed EMCM and QBC, and both MT-iGS and MT-GSy outperformed their single-task counterparts. For a given $K$, the average performances were generally in the order of MT-iGS $>$ MT-GSy $>$ iGS $>$ GSy $>$ GSx $>$ EMCM $>$ QBC $>$ BL1.
\end{enumerate}

Similar observations can also be made from Figs.~\ref{fig:RR2} and \ref{fig:RR3}, except that in Fig.~\ref{fig:RR2} MT-GSy and GSy (MT-iGS and iGS) had comparable performances on Arousal estimation, and in Fig.~\ref{fig:RR3} MT-GSy and GSy (MT-iGS and iGS) had comparable performances on Dominance estimation.

To quantify the performance improvements of different ALR approaches over the random sampling approach (BL1), we picked $K=\{50,100,150,200,250\}$ and recorded the performances (RMSE and CC) of BL1, shown in the fourth column of Table~\ref{tab:imp}. Then, we also show the performances of the seven ALR approaches in Columns 5-11 of Table~\ref{tab:imp}, where the numbers in the parentheses representing the performance improvements over the corresponding BL1 performance, and the best two are marked in bold. For example,  the first row shows that for Valence, using 50 samples, BL1 achieved an RMSE of 0.380, whereas EMCM achieved an RMSE of 0.356, representing a $(0.380-0.356)/0.380=6\%$ improvement.

Table~\ref{tab:imp} shows that:
\begin{enumerate}
\item Given a specific $K$, all ALR approaches outperformed BL1 in both RMSE and CC. Among them, MT-GSy and MT-iGS almost always achieved the best performance.
\item As $K$ increased, the performance improvement of ALR approaches decreased. This is because when $K$ becomes larger, the samples selected by different approaches, no matter ALR or BL1, overlap more, and hence the performance differences among them become smaller.
\end{enumerate}
These observations are consistent with those made from Figs.~\ref{fig:RR1}-\ref{fig:RR3}.

\begin{table*}[htpb] \centering \setlength{\tabcolsep}{2mm}
\caption{Performances and percentages of improvement (in the parentheses) when different ALR approaches are compared with BL1. The best two in each row are marked in bold.}   \label{tab:imp}
\begin{tabular}{c|c|cc|ccccccc} \hline
Emotion  & Performance & \multirow{2}{*}{K}&\multirow{2}{*}{BL1} & \multicolumn{7}{|c}{Performance and percentage improvement over BL1}   \\
Primitive  & Measure &   & &  EMCM  & QBC  & GSx & GSy & iGS & MT-GSy& MT-iGS  \\ \hline
    \multirow{10}{*}{Valence} & \multirow{5}{*}{RMSE}& 50 & 0.380 & 0.356 (6\%) & 0.361 (5\%) & 0.326 (14\%) & 0.311 (18\%) & 0.310 (18\%) & \textbf{0.300 (21\%)} & \textbf{0.299 (21\%)} \\
     & & 100 & 0.252 & 0.235 (7\%) & 0.237 (6\%) & 0.237 (6\%) & 0.232 (8\%) & 0.230 (9\%) & \textbf{0.226 (10\%)} & \textbf{0.225 (11\%)} \\
     & & 150 & 0.226 & 0.217 (4\%) & 0.217 (4\%) & 0.219 (3\%) & 0.216 (4\%) & 0.216 (4\%) & \textbf{0.214 (5\%)} & \textbf{0.213 (6\%)} \\
     & & 200 & 0.213 & 0.210 (2\%) & 0.210 (2\%) & 0.210 (1\%) & 0.210 (2\%) & 0.210 (2\%) & \textbf{0.209 (2\%)} & \textbf{0.208 (2\%)} \\
     & & 250 & 0.207 & 0.206 (1\%) & 0.206 (1\%) & 0.206 (1\%) & 0.206 (1\%) &0.206 (1\%) & \textbf{0.206 (1\%)} & \textbf{0.205 (1\%)} \\ \cline{2-11}
     & \multirow{5}{*}{CC}& 50 & 0.354 & 0.371 (5\%) & 0.367 (4\%) & 0.424 (20\%) & 0.434 (23\%) & 0.437 (23\%) & \textbf{0.446 (26\%)} & \textbf{0.448 (26\%)} \\
     & & 100 & 0.529 & 0.560 (6\%) & 0.553 (5\%) & 0.560 (6\%) & 0.561 (6\%) & 0.568 (7\%) & \textbf{0.574 (8\%)} & \textbf{0.579 (9\%)} \\
     & & 150 & 0.581 & 0.604 (4\%) & 0.600 (3\%) & 0.597 (3\%) & 0.599 (3\%) & 0.603 (4\%) & \textbf{0.606 (4\%)} & \textbf{0.609 (5\%)} \\
     & & 200 & 0.610 & 0.621 (2\%) & 0.619 (1\%) & 0.618 (1\%) & 0.618 (1\%) & 0.619 (1\%) & \textbf{0.622 (2\%)} & \textbf{0.623 (2\%)} \\
     & & 250 & 0.626 & 0.630 (1\%) & 0.630 (1\%) & 0.629 (1\%) & 0.630 (1\%) & \textbf{0.631 (1\%)} & 0.630 (1\%) & \textbf{0.631 (1\%)} \\ \hline
    \multirow{10}{*}{Arousal} & \multirow{5}{*}{RMSE}& 50 & 0.374 & 0.350 (6\%) & 0.357 (4\%) & 0.330 (12\%) & 0.311 (17\%) & 0.308 (18\%) & \textbf{0.300 (20\%)} & \textbf{0.298 (20\%)} \\
     & & 100 & 0.253 & 0.235 (7\%) & 0.236 (7\%) & 0.234 (7\%) & 0.235 (7\%) & 0.232 (8\%) & \textbf{0.226 (11\%)} & \textbf{0.225 (11\%)} \\
     & & 150 & 0.224 & 0.217 (3\%) & 0.216 (4\%) & 0.216 (4\%) & 0.219 (2\%) & 0.217 (3\%) & \textbf{0.213 (5\%)} & \textbf{0.213 (5\%)} \\
     & & 200 & 0.213 & 0.209 (2\%) & 0.209 (2\%) & 0.209 (2\%) & 0.210 (1\%) & 0.209 (2\%) & \textbf{0.208 (3\%)} & \textbf{0.208 (3\%)} \\
     & & 250 & 0.207 & 0.205 (1\%) & 0.205 (1\%) & 0.205 (1\%) & 0.206 (1\%) & 0.205 (1\%) & \textbf{0.205 (1\%)} & \textbf{0.205 (1\%)} \\ \cline{2-11}
     & \multirow{5}{*}{CC}& 50 & 0.368 & 0.393 (7\%) & 0.379 (3\%) & 0.419 (14\%) & 0.436 (18\%) & 0.442 (20\%) & \textbf{0.447 (21\%)} & \textbf{0.449 (22\%)} \\
     & & 100 & 0.529 & 0.559 (6\%) & 0.554 (5\%) & 0.567 (7\%) & 0.557 (5\%) & 0.564 (7\%) & \textbf{0.573 (8\%)} & \textbf{0.576 (9\%)} \\
     & & 150 & 0.584 & 0.603 (3\%) & 0.599 (3\%) & 0.604 (3\%) & 0.593 (1\%) & 0.600 (3\%) & \textbf{0.606 (4\%)} & \textbf{0.608 (4\%)} \\
     & & 200 & 0.609 & 0.620 (2\%) & 0.620 (2\%) & 0.621 (2\%) & 0.615 (1\%) & 0.619 (2\%) & \textbf{0.622 (2\%)} & \textbf{0.622 (2\%)} \\
     & & 250 & 0.626 & 0.630 (1\%) & 0.630 (1\%) & 0.630 (1\%) & 0.628 (0\%) & 0.630 (1\%) & \textbf{0.631 (1\%)} & \textbf{0.631 (1\%)} \\ \hline
    \multirow{10}{*}{Dominance} & \multirow{5}{*}{RMSE} & 50 & 0.370 & 0.354 (4\%) & 0.359 (3\%) & 0.321 (13\%) & 0.304 (18\%) & 0.303 (18\%) & \textbf{0.296 (20\%)} & \textbf{0.296 (20\%)} \\
     & & 100 & 0.251 & 0.236 (6\%) & 0.235 (6\%) & 0.235 (7\%) & 0.233 (7\%) & 0.231 (8\%) & \textbf{0.224 (11\%)} & \textbf{0.224 (11\%)} \\
     & & 150 & 0.224 & 0.217 (3\%) & 0.217 (3\%) & 0.217 (3\%) & 0.217 (3\%) & 0.216 (4\%) & \textbf{0.213 (5\%)} & \textbf{0.213 (5\%)} \\
     & & 200 & 0.213 & 0.209 (2\%) & 0.209 (2\%) & 0.209 (2\%) & 0.210 (2\%) & 0.210 (1\%) & \textbf{0.208 (2\%)} & \textbf{0.208 (2\%)} \\
     & & 250 & 0.207 & 0.205 (1\%) &0.205 (1\%) & 0.205 (1\%) & 0.205 (1\%) & 0.206 (1\%) & \textbf{0.205 (1\%)} & \textbf{0.205 (1\%)} \\ \cline{2-11}
     & \multirow{5}{*}{CC} & 50 & 0.377 & 0.388 (3\%) & 0.384 (2\%) & 0.433 (15\%) & 0.445 (18\%) & 0.446 (18\%) & \textbf{0.453 (20\%)} & \textbf{0.454 (21\%)} \\
     & & 100 & 0.536 & 0.560 (5\%) & 0.559 (4\%) & 0.566 (6\%) & 0.558 (4\%) & 0.566 (6\%) & \textbf{0.579 (8\%)} & \textbf{0.579 (8\%)} \\
     & & 150 & 0.586 & 0.602 (3\%) & 0.600 (2\%) & 0.601 (3\%) & 0.597 (2\%) & 0.601 (3\%) & \textbf{0.607 (4\%)} & \textbf{0.608 (4\%)} \\
     & & 200 & 0.611 &0.621 (2\%) & 0.619 (1\%) & 0.620 (2\%) & 0.616 (1\%) & 0.617 (1\%) & \textbf{0.621 (2\%)} & \textbf{0.623 (2\%)} \\
     & & 250 & 0.626 &0.630 (1\%) & 0.630 (1\%) & 0.630 (1\%) & 0.629 (1\%) & 0.629 (1\%) & \textbf{0.630 (1\%)} & \textbf{0.631 (1\%)} \\ \hline
\end{tabular}
\end{table*}

\subsection{ALR Saved the Number of Queries over BL1}

The improved performances of ALR can also be verified by quantifying the numbers of saved queries over BL1, when a desired regression performance is needed.

To do this, we first count the number of labeled samples required for BL1 to achieve $(100+\alpha)\%$ RMSE of BL2 ($\alpha=\{1,2,3,5,10\}$), and $(100-\alpha)\%$ CC of BL2, as shown in the fourth column of Table~\ref{tab:saved}. We then also count the number of samples required by different ALR approaches, and the corresponding saving over BL1, as shown in the remaining columns of Table~\ref{tab:saved}. For example, the first row shows that for Valence, to achieve 101\% ($\alpha=1$) RMSE of BL2, BL1 needed 261 labeled samples, whereas EMCM only needed 242 samples, representing a $(261-242)/261=8\%$ saving.

Table~\ref{tab:saved} shows that:
\begin{enumerate}
\item All ALR approaches can save the number of queries over BL1. Among them, MT-GSy and MT-iGS, particularly MT-iGS, almost always saved the most number of queries.
\item As $\alpha$ increased, the percentage of saving also increased for almost all ALR approaches, especially MT-GSy and MT-iGS.
\end{enumerate}
These observations are consistent with those made in the previous subsection.

\begin{table*}[htpb] \centering \setlength{\tabcolsep}{2mm}
\caption{Number of samples and percentages of saved queries (in the parentheses) when different ALR approaches are compared with BL1. The best two in each row are marked in bold.}   \label{tab:saved}
\begin{tabular}{c|c|cc|ccccccc} \hline
Emotion  & Performance & \multirow{2}{*}{$\alpha\%$}   & No. BL1 & \multicolumn{7}{|c}{Number of samples and percentage saving over BL1}   \\
Primitive  & Measure &  & Samples &  EMCM  & QBC  & GSx & GSy & iGS & MT-GSy& MT-iGS  \\ \hline
    \multirow{10}{*}{Valence} & \multirow{5}{*}{RMSE}& 1\% & 261 & \textbf{242 (8\%)} & 248 (5\%) & 247 (6\%) & 246 (6\%) & 242 (8\%) & 243 (7\%) & \textbf{233 (12\%)} \\
     & & 2\% & 241 & 218 (11\%) & 217 (11\%) & 221 (9\%) & 218 (11\%) & 216 (12\%) & \textbf{207 (16\%)} & \textbf{202 (19\%)} \\
     & & 3\% & 222 & 197 (13\%) & 197 (13\%) & 201 (10\%) & 194 (14\%) & 197 (13\%) & \textbf{183 (21\%)} & \textbf{179 (24\%)} \\
     & & 5\% & 197 & 168 (17\%) & 168 (17\%) & 175 (13\%) & 164 (20\%) & 162 (22\%) & \textbf{148 (33\%)} & \textbf{144 (37\%)} \\
     & & 10\% & 154 & 123 (25\%) & 126 (22\%) & 129 (19\%) & 118 (31\%) & 116 (33\%) & \textbf{106 (45\%)} & \textbf{101 (52\%)} \\ \cline{2-11}
     & \multirow{5}{*}{CC}& 1\% & 258 & \textbf{236 (9\%)} & 242 (7\%) & 242 (7\%) & 244 (6\%) & 238 (8\%) & 242 (7\%) & \textbf{230 (12\%)} \\
     & & 2\% & 235 & 202 (16\%) & 211 (11\%) & 211 (11\%) & 215 (9\%) & 208 (13\%) & \textbf{201 (17\%)} & \textbf{194 (21\%)} \\
     & & 3\% & 215 & 181 (19\%) & 187 (15\%) & 190 (13\%) & 189 (14\%) & 185 (16\%) & \textbf{175 (23\%)} & \textbf{172 (25\%)} \\
     & & 5\% & 184 & 149 (23\%) & 154 (19\%) & 162 (14\%) & 157 (17\%) & 150 (23\%) & \textbf{144 (28\%)} & \textbf{135 (36\%)} \\
     & & 10\% & 138 & 109 (27\%) & 115 (20\%) & 112 (23\%) & 110 (25\%) & 104 (33\%) & \textbf{98 (41\%)} & \textbf{93 (48\%)} \\ \hline
    \multirow{10}{*}{Arousal} & \multirow{5}{*}{RMSE}& 1\% & 261 & 239 (9\%) & \textbf{237 (10\%)} & 245 (7\%) & 252 (4\%) & 247 (6\%) & \textbf{231 (13\%)} & 238 (10\%) \\
     & & 2\% & 242 & 213 (14\%) & 210 (15\%) & 216 (12\%) & 227 (7\%) & 220 (10\%) & \textbf{199 (22\%)} & \textbf{196 (23\%)} \\
     & & 3\% & 225 & 193 (17\%) & 192 (17\%) & 196 (15\%) & 208 (8\%) & 197 (14\%) & \textbf{176 (28\%)} & \textbf{174 (29\%)} \\
     & & 5\% & 196 & 165 (19\%) & 165 (19\%) & 167 (17\%) & 175 (12\%) & 161 (22\%) & \textbf{143 (37\%)} & \textbf{139 (41\%)} \\
     & & 10\% & 152 & 125 (22\%) & 125 (22\%) & 126 (21\%) & 126 (21\%) & 116 (31\%) & \textbf{98 (55\%)} & \textbf{97 (57\%)} \\ \cline{2-11}
     & \multirow{5}{*}{CC}& 1\% & 261 & 235 (11\%) & 235 (11\%) & 242 (8\%) & 247 (6\%) & 241 (8\%) & \textbf{231 (13\%)} & \textbf{228 (14\%)} \\
     & & 2\% & 241 & 202 (19\%) & 203 (19\%) & 208 (16\%) & 219 (10\%) & 210 (15\%) & \textbf{198 (22\%)} & \textbf{188 (28\%)} \\
     & & 3\% & 223 & 181 (23\%) & 184 (21\%) & 186 (20\%) & 200 (12\%) & 188 (19\%) & \textbf{174 (28\%)} & \textbf{165 (35\%)} \\
     & & 5\% & 185 & 147 (26\%) & 155 (19\%) & 155 (19\%) & 168 (10\%) & 152 (22\%) & \textbf{135 (37\%)} & \textbf{132 (40\%)} \\
     & & 10\% & 136 & 110 (24\%) & 113 (20\%) & 105 (30\%) & 115 (18\%) & 105 (30\%) & \textbf{92 (48\%)} & \textbf{91 (49\%)} \\ \hline
    \multirow{10}{*}{Dominance} & \multirow{5}{*}{RMSE} & 1\% & 261 & 237 (10\%) & 238 (10\%) & 237 (10\%) & 249 (5\%) & 242 (8\%) & \textbf{232 (13\%)} & \textbf{235 (11\%)} \\
     & & 2\% & 242 & 210 (15\%) & 212 (14\%) & 213 (14\%) & 221 (10\%) & 216 (12\%) & \textbf{204 (19\%)} & \textbf{201 (20\%)} \\
     & & 3\% & 225 & 192 (17\%) & 194 (16\%) & 191 (18\%) & 204 (10\%) & 193 (17\%) & \textbf{182 (24\%)} & \textbf{180 (25\%)} \\
     & & 5\% & 197 & 165 (19\%) & 166 (19\%) & 162 (22\%) & 174 (13\%) & 164 (20\%) & \textbf{148 (33\%)} & \textbf{146 (35\%)} \\
     & & 10\% & 151 & 123 (23\%) & 127 (19\%) & 124 (22\%) & 126 (20\%) & 121 (25\%) & \textbf{106 (42\%)} & \textbf{103 (47\%)} \\ \cline{2-11}
     & \multirow{5}{*}{CC} & 1\% & 258 & 236 (9\%) & 239 (8\%) & 231 (12\%) & 248 (4\%) & 238 (8\%) & \textbf{226 (14\%)} & \textbf{229 (13\%)} \\
     & & 2\% & 234 & 200 (17\%) & 210 (11\%) & 203 (15\%) & 219 (7\%) & 211 (11\%) & \textbf{196 (19\%)} & \textbf{196 (19\%)} \\
     & & 3\% & 217 & 181 (20\%) & 190 (14\%) & 181 (20\%) & 197 (10\%) & 186 (17\%) & \textbf{175 (24\%)} & \textbf{173 (25\%)} \\
     & & 5\% & 184 & 148 (24\%) & 159 (16\%) & 147 (25\%) & 167 (10\%) & 155 (19\%) & \textbf{143 (29\%)} & \textbf{139 (32\%)} \\
     & & 10\% & 133 & 110 (21\%) & 116 (15\%) & 104 (28\%) & 113 (18\%) & 107 (24\%) & \textbf{96 (39\%)} & \textbf{95 (40\%)} \\ \hline
\end{tabular}
\end{table*}

\subsection{Model Parameters from ALR Converged Faster}

\begin{figure*}[htpb]\centering
\includegraphics[width=.96\linewidth,clip]{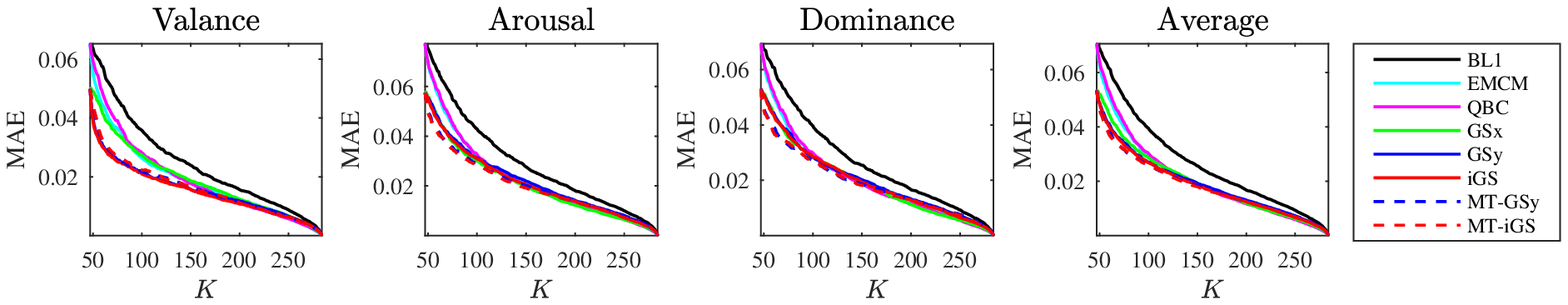}
\caption{MAE between the regression coefficients of BL2 and the other eight approaches, when the single-task ALR approaches focused on \emph{Valence}. RR was used as the regression model.} \label{fig:Para}
\end{figure*}

As BL2 used all samples in the training pool, the regression coefficients obtained from BL2 represented the global optimum. It's interesting to study how fast the model parameters from different approaches converged to the solution given by BL2. The mean absolute errors (MAEs) between the coefficients of BL2 and the other eight approaches for different $K$ are shown in Fig.~\ref{fig:Para}. To save space, we only show the results when single-task ALR approaches focused on \emph{Valence}.

Fig.~\ref{fig:Para} shows that:
\begin{enumerate}
\item Generally as $K$ increased, the model parameters from all eight sample selection algorithms converged to the solution of BL2.
\item When $K=K_0=46$ (the first point in each subfigure of Fig.~\ref{fig:Para}), the five GS based ALR approaches (GSx, GSy, iGS, MT-GSy and MT-iGS), which initialized the $K_0$ samples by considering the diversity in the input space, all had smaller MAEs than the other three approaches (BL, EMCM and QBC), which initialized the $K_0$ samples randomly.
\item GSy and iGS achieved comparable MAEs with MT-GSy and MT-iGS on Valence estimation, because this was the task that GSy and iGS focused on; however, for the other two tasks (Arousal and Dominance estimations), MT-GSy and MT-iGS achieved smaller MAEs.
\item When all three tasks are considered together (the last column of Fig.~\ref{fig:Para}), generally all ALR approaches converged faster than BL1, all GS based approaches converged faster than EMCM and QBC, and both MT-iGS and MT-GSy converged faster than their single-task counterparts.
\end{enumerate}

\subsection{Sample Selection Results: Impact of Gender}

\begin{figure}[htpb]\centering
\includegraphics[width=.8\linewidth,clip]{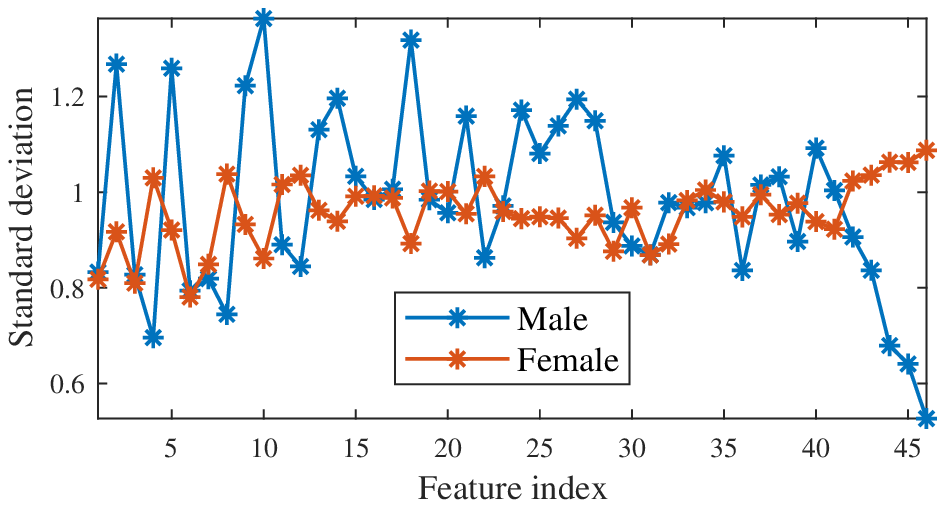}
\caption{Feature standard deviations of male and female utterances.} \label{fig:stdGender}
\end{figure}

The VAM dataset consists of 947 utterances, 196 of which are from males ($20.7\%$), and 751 from females ($79.3\%$). Male and female utterances have different feature standard deviations, as shown in Fig.~\ref{fig:stdGender}. For 27 of the 46 features, male utterances have larger standard deviations than female utterances.

\begin{figure*}[htpb]\centering
\includegraphics[width=.8\linewidth,clip]{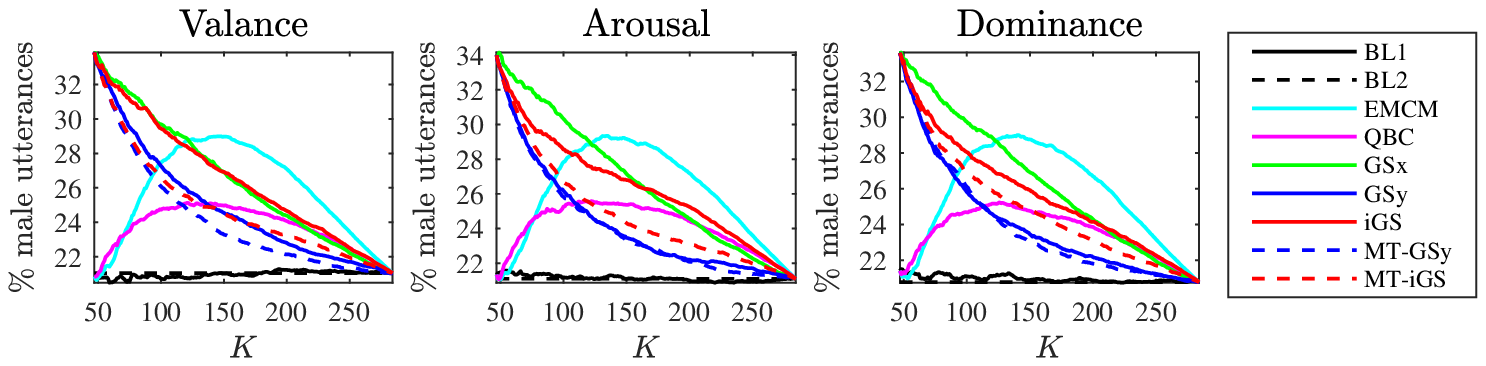}
\caption{Percentage of male utterances selected by different algorithms, when the single-task ALR approaches focused on \emph{Valence}, \emph{Arousal}, and \emph{Dominance}, respectively. RR was used as the regression model.} \label{fig:pMale}
\end{figure*}

As the initial 46 samples from the GS based ALR approaches are selected based on the diversity in the feature space, and the male utterances have larger feature variations than the female utterances, we expect that more male utterances would be selected by the GS based ALR approaches in initialization than other approaches. Fig.~\ref{fig:pMale}, which shows the percentage of male utterances selected by different algorithms, confirms this. When $K=46$, BL1, EMCM and QBC selected about $20.7\%$ male utterances, which is the average percentage of male utterances in the dataset. However, all five GS based ALR approaches selected over $33\%$ male utterances, much larger than the average percentage. As $K$ increased, this percentage gradually decreased. Interestingly, the percentage of male utterances selected by EMCM and QBC first increased with $K$, and then gradually decreased.

\subsection{Sample Selection Results: Impact of Emotion Primitive Values}

\begin{figure*}[htpb]\centering
\includegraphics[width=\linewidth,clip]{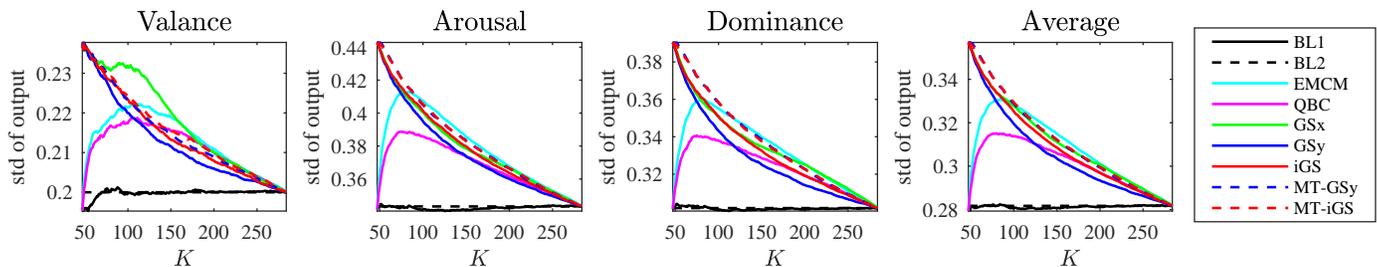}
\caption{Standard deviation (std) of the primitives of the selected samples, when the single-task ALR approaches focused on \emph{Valence}. RR was used as the regression model.} \label{fig:stdY}
\end{figure*}

To study how the values of the emotion primitives impacted the sample selection algorithms, we computed the standard deviation of the primitives of the selected samples, and show the results in Fig.~\ref{fig:stdY}, when the single-task ALR approaches focused on Valence. The samples selected by the five GS based ALR approaches generally had larger standard deviations than those by other approaches, especially when $K$ was small. This is intuitive, as GS tends to select the most diverse samples. The first $K_0$ samples selected by EMCM and QBC had the same standard deviation as those by BL1, as all of them were random. However, as $K$ increased, the standard deviation of the samples selected by EMCM and QBC increased rapidly and were much larger than that by BL1. The standard deviation of the samples selected by EMCM was larger than that by QBC, and the performance of EMCM was also slightly better than QBC (Table~\ref{tab:imp}).

\subsection{Impact of Features on Algorithm Performance}

\begin{figure}[htpb]\centering
\includegraphics[width=\linewidth,clip]{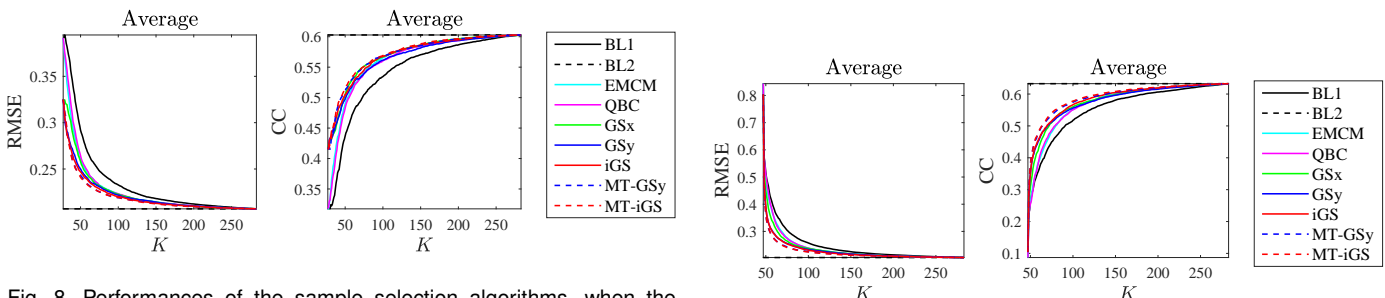}
\caption{Performances of the sample selection algorithms, when the single-task ALR approaches focused on \emph{Valence}. RR was used as the regression model, and only the 26 MFCC features were used.} \label{fig:MFCC}
\end{figure}

This subsection evaluates the performances of different ALR approaches w.r.t. different feature sets. We repeated the experiments in Section~\ref{sect:RR} using only the 26 MFCC features instead of all 46 features. To save space, here we only show the average results across the three tasks in Fig.~\ref{fig:MFCC}, when the single-task ALR approaches focused on Valence. The average performances of the algorithms are in the order of MT-iGS $>$ MT-GSy $>$ iGS $>$ GSy $>$ GSx $>$ EMCM $>$ QBC $>$ BL1. The RMSEs in Fig.~\ref{fig:MFCC} are similar to those in Fig.~\ref{fig:RR1}, but the CCs were generally smaller, i.e., reducing the number of features did not change the rank of the algorithms, but resulted in overall worse performances for all approaches.

\subsection{Impact of Regression Models on Algorithm Performance}

This subsection studies the stability of the proposed multi-task ALR approaches w.r.t. different regression models. To save space,  we only present the results when the single-task ALR approaches focused on Valence.

\begin{figure}[htpb]\centering
\subfigure[]{\label{fig:RLS}     \includegraphics[width=\linewidth,clip]{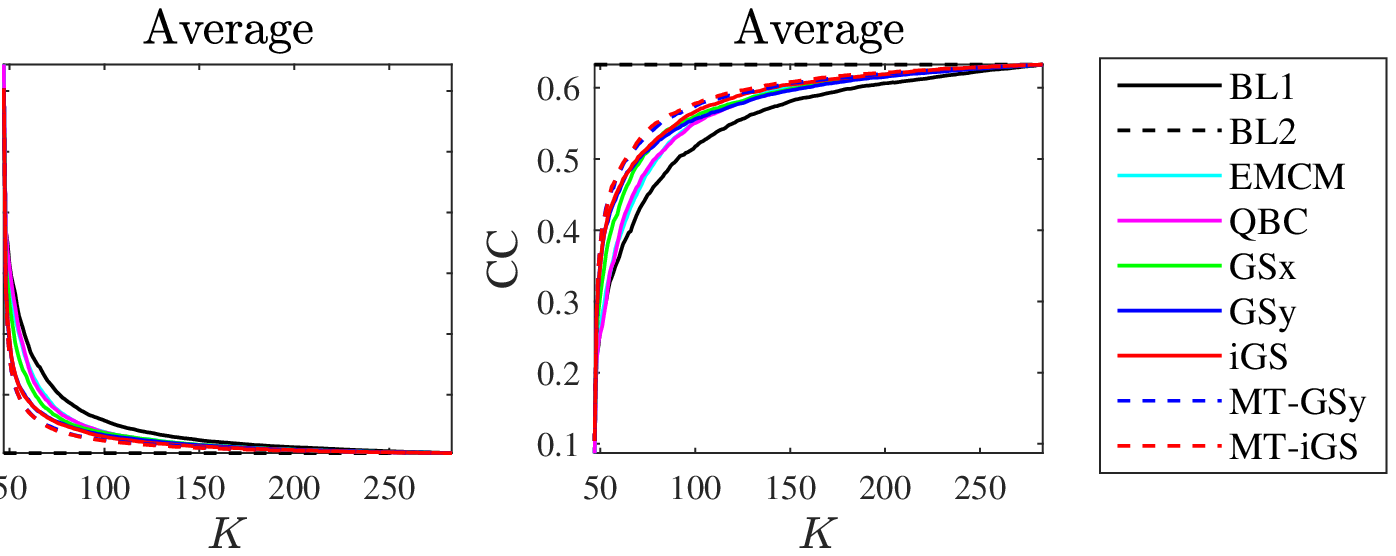}}
\subfigure[]{\label{fig:LASSO}     \includegraphics[width=\linewidth,clip]{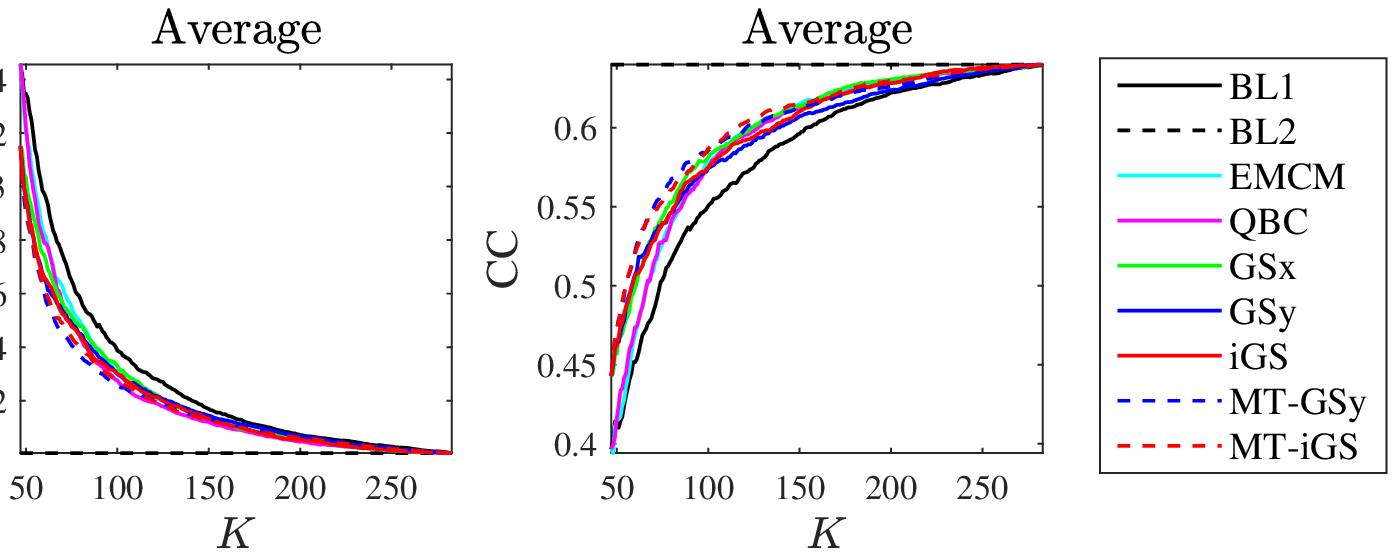}}
\subfigure[]{\label{fig:EN}     \includegraphics[width=\linewidth,clip]{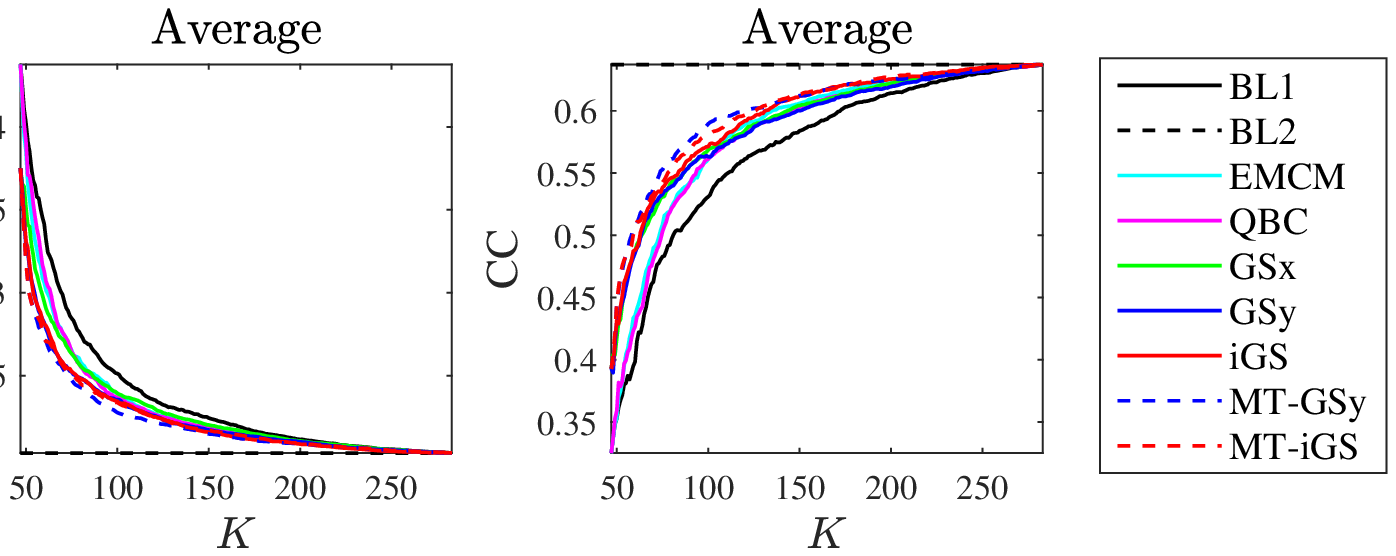}}
\caption{Performances of the sample selection algorithms, when the single-task ALR approaches focused on \emph{Valence}. Different regression models were used. (a) Ordinary least square regression; (b) LASSO; (c) Elastic net.} \label{fig:algs}
\end{figure}

First, ordinary least square regression was used as the linear regression model. We ran the experiments as previously for RR, where the single-task ALR approaches focused on different tasks. The results are shown in Fig.~\ref{fig:RLS}. Again we can make similar observations as those for RR, although initially ordinary least square regression performed much worse than RR, as there were not enough training samples to adequately determine the coefficients of the features, and no regularization was done either on the coefficients.

Next, LASSO was used as the linear regression model, where $\lambda=0.001$ in its objective function $ \underset{\boldsymbol{\beta}}{\operatorname{min}} (\| \mathbf{y}-\mathbf{X} \boldsymbol{\beta} \|^2 + \lambda \|\boldsymbol{\beta}\|_1)$. The average results are shown in Fig.~\ref{fig:LASSO}. Finally, elastic net was used as the linear regression model, where $\lambda_1=\lambda_2=0.0005$ in its objective function $\underset{\boldsymbol{\beta}}{\operatorname{min}} (\| \mathbf{y}-\mathbf{X} \boldsymbol{\beta} \|^2 + \lambda_1 \|\boldsymbol{\beta}\|_1) + \lambda_2 \|\boldsymbol{\beta}\|^2$. The average results are shown in Fig.~\ref{fig:EN}. For a given $K$, the performances were in the order of MT-iGS $\approx$ MT-GSy $>$ iGS $>$ GSy $>$ GSx $>$ EMCM $>$ QBC $>$ BL, consistent with those when RR was used as the regression model.

\section{Discussion: Why MTL in Affective Computing} \label{sect:discussion}

The previous section has presented extensive experiments and comprehensive analyses, showing that our proposed multi-task ALR approaches, MT-GSy and MT-iGS, outperformed their single-task counterparts, GSy and iGS, and also three other state-of-the-art single-task ALR approaches in the literature. However, a natural question that a user may ask is: Why should MTL be used in affective computing at the first place, instead of viewing each emotion primitive estimation as a separate problem and acquiring the labeled samples completely independently?

Our argument is that usually it takes time to evaluate (label) an affective signal, whether it is text, image, utterance, video, or others. So, a multi-task labeling approach, i.e., evaluating a single affective signal and then assigning valence, arousal and dominance simultaneously to it, is much more efficient than the combination of three separate single-task approaches, i.e., first evaluating Signal~1 and assigning Valence to it, then evaluating a different Signal~2 and assigning Arousal to it, and finally evaluating another different Signal~3 and assigning Dominance to it. This is particularly true when the affective signals are long and time-consuming to evaluate, e.g., movies. To obtain a label each for valence, arousal and dominance, the multi-task approach requires the assessor to watch only one movie and then assign three primitive values, whereas the single-task approach needs the assessor to watch three different movies. The former is usually much faster, easier, and more user-friendly. A well-designed MTL algorithm, like MT-GSy or MT-iGS, lets one select a small number of most informative affective signals to label, and can achieve comparable performance as the combination of three optimal single-task algorithms, while saving a significant amount of evaluation time.

\begin{figure}[htpb]\centering
\includegraphics[width=.7\linewidth,clip]{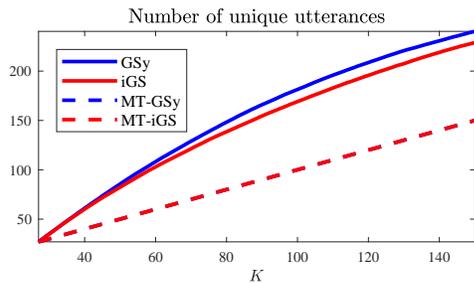}
\caption{Number of unique utterances when each single-task ALR approach (GSy and iGS) was independently optimized for the three emotion primitives. RR was used as the regression model, and only the 26 MFCC features were used. The curves for MT-GSy and MT-iGS are identical.} \label{fig:MTSTnum}
\end{figure}

As an example, we performed an experiment using RR and 26 MFCC features on the VAM dataset. A random 30\% pool was reserved for training, and the remaining 70\% for testing. $K$ increased from 26 to 150. The first 26 samples for GSy, iGS, MT-GSy and MT-iGS were selected using GSy and were the same for all four approaches. Then, MT-GSy and MT-iGS proceeded just as before, but we used GSy (iGS) to separately select the optimal samples for Valence, Arousal and Dominance. For each $K$, we counted the number of unique utterances selected by the multi-task and single-task approaches. For example, when $K=50$, MT-GSy and MT-iGS each selected 50 unique utterances, but separately running GSy on Valence, Arousal and Dominance estimations selected 150 utterances, of which 87 were unique. The results are shown in Fig.~\ref{fig:MTSTnum}. Clearly, the single-task ALR approaches used much more unique utterances than the multi-task ones.

\begin{figure*}[htpb]\centering
\includegraphics[width=.9\linewidth,clip]{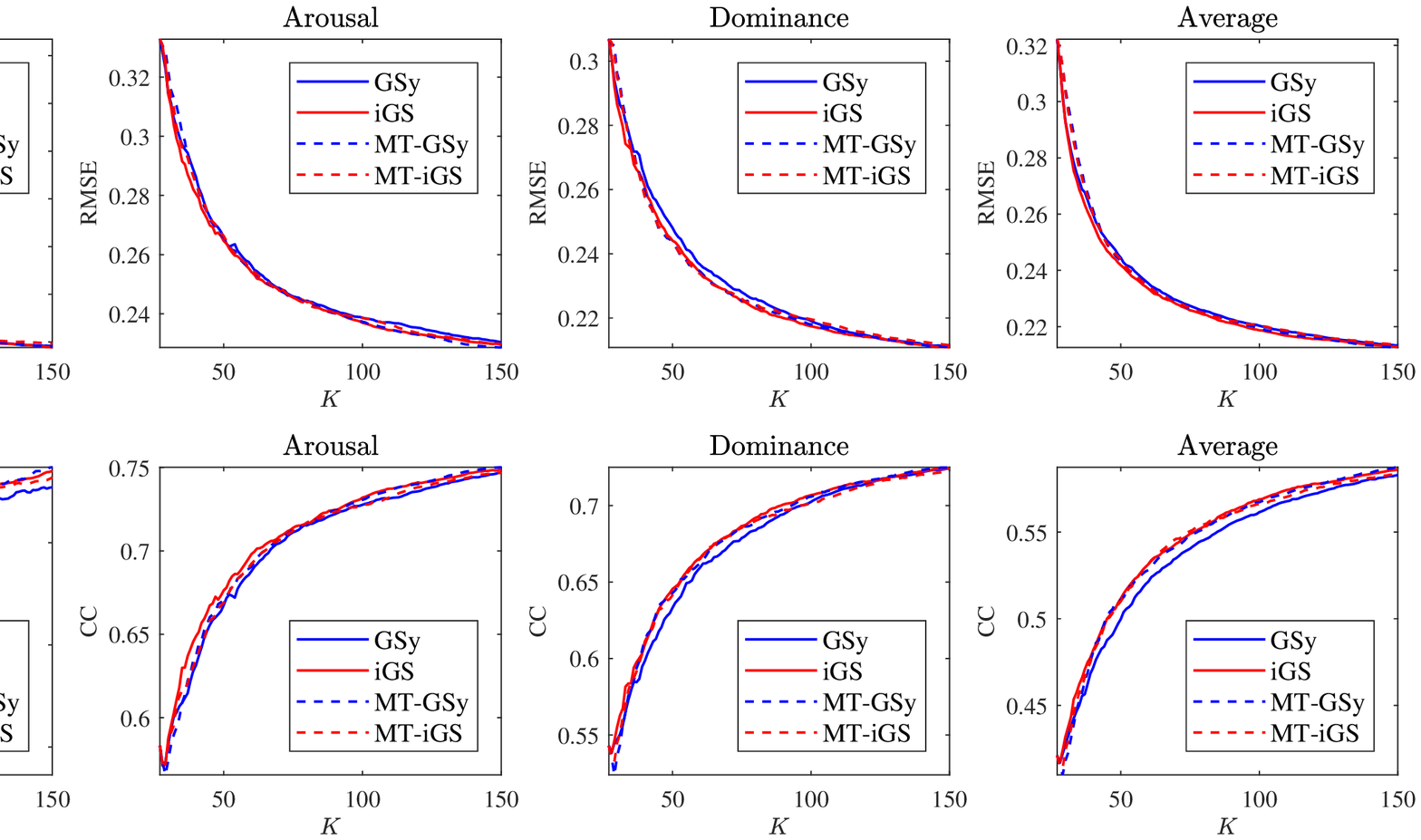}
\caption{Estimation performances when each single-task ALR approach (GSy and iGS) was independently optimized for the three emotion primitives. RR was used as the regression model, and only the 26 MFCC features were used.} \label{fig:MTST}
\end{figure*}

Fig.~\ref{fig:MTST} shows the estimation performances of the multi-task and single-task approaches, based on the samples selected in Fig.~\ref{fig:MTSTnum}. MT-iGS and MT-GSy achieved similar performances as three independent single-task GSy or iGS combined.

In summary, MTL can achieve similar estimation performances as optimizing multiple single-tasks separately, but can significantly reduce the number of unique utterances that an assessor needs to evaluate. So, MTL is advantageous and effective in affective computing.

\section{Conclusions and Future Research} \label{sect:conclusions}

Acquisition of labeled training samples for affective computing is usually costly and time-consuming, as multiple human assessors are needed to evaluate each affective sample. Particularly, for affect estimation in the 3D space of valence, arousal and dominance, each assessor has to perform the evaluations in three dimensions, which makes the labeling problem even more challenging. This paper has proposed two MT-ALR approaches, MT-GSy and MT-iGS, which select the most informative samples to label, by considering the three affect primitives simultaneously. Experimental results on the VAM corpus demonstrated that MT-GSy and MT-iGS outperformed random selection and several traditional single-task ALR approaches, i.e., better affect estimation performance can be achieved when MT-GSy or MT-iGS is used to select the affective samples to label. Or, in other words, when a desired performance (RMSE or CC) is needed, using MT-GSy or MT-iGS can save the number of labelling queries.

Our future research directions include:
\begin{enumerate}
\item Extend MT-GSy and MT-iGS from regression to classification, as affects can also be classified simultaneously in multiple dimensions, e.g., paralinguistics in speech and language \cite{Schuller2013a}.

\item Extend our MT-ALR approaches from offline pool-based regression to online streaming regression \cite{Settles2009}.

\item Develop new MT-ALR approaches for nonlinear regression models, e.g., deep neural networks, as \cite{drwuiGS2018} shows that GSy and iGS do not perform well when nonlinear regression models are used.

\item Consider the more general case that different tasks use different inputs, in contrast to the case in this paper that all tasks share the same inputs.
\end{enumerate}

%\section*{Acknowledgement}
%This research was supported by the National Natural Science Foundation of China (61873321).

%\bibliographystyle{IEEETranS}\bibliography{drwubib}

% Generated by IEEEtranS.bst, version: 1.14 (2015/08/26)

\end{document}